\documentclass[conference]{IEEEtran}
\IEEEoverridecommandlockouts
\usepackage{cite}
\usepackage{amsmath,amssymb,amsfonts}
\usepackage{algorithmic}
\usepackage{graphicx}
\usepackage{textcomp}
\usepackage{xcolor}
\usepackage{multirow}

\usepackage{hyperref}

\hypersetup{
    colorlinks=true,
    linkcolor=blue,
    filecolor=magenta,      
    urlcolor=cyan,
    citecolor=blue,
}
\def\BibTeX{{\rm B\kern-.05em{\sc i\kern-.025em b}\kern-.08em
    T\kern-.1667em\lower.7ex\hbox{E}\kern-.125emX}}
\begin{document}

\title{DCG-Net: Dual Cross-Attention with Concept-Value Graph Reasoning for Interpretable Medical Diagnosis}

\author{
\IEEEauthorblockN{
Getamesay Haile Dagnaw,
Xuefei Yin,
Muhammad Hassan Maqsood,
Yanming Zhu,
Alan Wee-Chung Liew*\thanks{*Corresponding author: a.liew@griffith.edu.au}
}
\IEEEauthorblockA{
\textit{School of Information and Communication Technology} \\
\textit{Griffith University} \\
QLD, Australia \\
\{getamesay.dagnaw, x.yin, muhammad.maqsood, yanming.zhu, a.liew\}@griffith.edu.au
}
}
\maketitle

\begin{abstract}
Deep learning models have achieved strong performance in medical image analysis, but their internal decision processes remain difficult to interpret. Concept Bottleneck Models (CBMs) partially address this limitation by structuring predictions through human-interpretable clinical concepts. However, existing CBMs typically overlook the contextual dependencies among concepts. To address these issues, we propose an end-to-end interpretable framework \emph{DCG-Net} that integrates multimodal alignment with structured concept reasoning. DCG-Net introduces a Dual Cross-Attention module that replaces cosine similarity matching with bidirectional attention between visual tokens and canonicalized textual concept-value prototypes, enabling spatially localized evidence attribution. To capture the relational structure inherent to clinical concepts, we develop a Parametric Concept Graph initialized with Positive Pointwise Mutual Information priors and refined through sparsity-controlled message passing. This formulation models inter-concept dependencies in a manner consistent with clinical domain knowledge. Experiments on white blood cell morphology and skin lesion diagnosis demonstrate that DCG-Net achieves state-of-the-art classification performance while producing clinically interpretable diagnostic explanations.

\end{abstract}


\section{Introduction}
\label{sec:intro}

Deep learning has achieved significant progress in medical image analysis, enabling advances in dermatologic screening, hematologic assessment, and other diagnostic pipelines~\cite{litjens2017survey, suganyadevi2022review, zhu2020neural}. However, the decision processes of deep learning models remain opaque, limiting their acceptability in high-stakes clinical environments~\cite{patricio2023explainable}. Saliency-based explanations alone are insufficient, as they highlight image regions without explaining the clinical rationale underlying these highlighted regions.

Concept Bottleneck Models (CBMs) provide a promising framework for interpretable medical image analysis by predicting human-interpretable concepts prior to diagnosis~\cite{koh2020concept}. However, classical CBMs assume concept-level independence, failing to capture contextual relationships among concepts. For example, dermatologic findings such as erythema, scale, and border irregularity frequently co-occur in clinically relevant co-occurrence structures~\cite{daneshjou2022skincon}. Ignoring such relationships limits the reliability of concept-based explanations and may yield inconsistent or clinically implausible reasoning.

Recent CBM extensions incorporate attention mechanisms, disentangled representations, or post-hoc concept extraction~\cite{kim2018interpretability,ghorbani2019towards,fang2020concept,yuksekgonul2022post,liu2025hybrid}, improving concept supervision but largely treating concepts independently during reasoning. Relational CBM variants, graph-based formulations, and concept-induced graph approaches~\cite{barbiero2024relational,xu2025graph,kim2024concept,zhao2025concept} address this limitation by explicitly modeling interactions among concepts, while neural-symbolic approaches~\cite{barbiero2023interpretable,gao2025learning} further couple learned representations with logical constraints to improve consistency.
However, these methods primarily model relationships at the concept level and overlook interactions between specific concept values. In clinical practice, diagnostic reasoning is often driven by value-level dependencies rather than concept presence alone; for example, in hematology, larger cells combined with a lower nuclear-cytoplasmic ratio suggest monocytes, whereas smaller cells with a higher ratio indicate lymphocytes~\cite{tsutsui2023wbcatt}. Modeling only concept-concept relations therefore risks missing the underlying clinical logic. Motivated by this limitation, we aim to develop a unified architecture that supports three capabilities essential for clinically aligned interpretability: 1) stable textual normalization of heterogeneous clinical concept definitions, 2) precise bidirectional cross-modal grounding that aligns concepts with visual evidence, and 3) structured reasoning that explicitly models concept-value dependencies.

To this end, we propose \textbf{DCG-Net}, a Dual Cross-Attention Network with Parametric Concept Graph reasoning. DCG-Net constructs canonical textual concept-value prototypes, aligns them with image representations through bidirectional attention, and refines concept activations using a clinically informed concept graph initialized with Positive Pointwise Mutual Information (PPMI) priors. This design allows the network to operate directly on concept-value pair representations (e.g., `Scale: present' vs.\ `erythema: present'), while capturing dependencies among concept values in realistic diagnostic scenarios. Experiments on skin lesion diagnosis~\cite{groh2021evaluating, daneshjou2022skincon} and white blood cell morphology classification~\cite{acevedo2020dataset,tsutsui2023wbcatt} show that DCG-Net produces interpretable predictions that closely align with expert diagnostic reasoning.

The main contributions are summarized as follows:
\begin{itemize}
    \item We propose an end-to-end concept bottleneck architecture \textbf{DCG-Net} that integrates vision-language pretraining with dual cross-attention and graph-based relational reasoning, ensuring that all diagnostic supervision is mediated through interpretable concept-value representations.
    \item We design a \emph{Dual Cross-Attention (DCA)} module that enables fine-grained bidirectional alignment between visual tokens and canonicalized textual prototypes, producing spatially grounded and semantically calibrated concept activations.
    \item We develop a \emph{Parametric Concept Graph (PCG)}, initialized with PPMI-based structural priors and refined through sparsity-controlled message passing, enabling clinically coherent modeling of concept dependencies.
    \item Extensive experiments on white blood cell morphology and skin lesion diagnosis demonstrate that DCG-Net achieves competitive diagnostic performance while substantially improving concept-level interpretability.
\end{itemize}

\section{Methodology}
\label{sec:method}

\begin{figure*}[t]
    \centering
    \includegraphics[width=\linewidth]{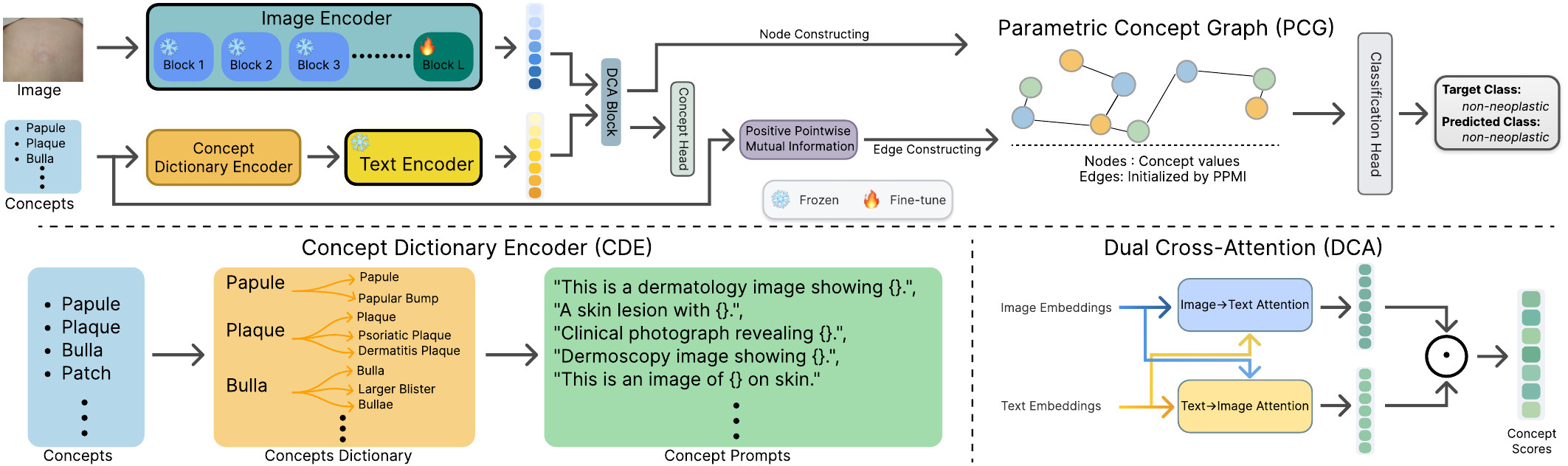} 
    \caption{
    \textbf{Overview of DCG-Net.}
    The \emph{CDE} generates \textbf{canonical concept-value prototypes} ($T_M$).
    A vision transformer encodes images into patch-level visual tokens ($V$).
    The \emph{DCA} module aligns visual and textual representations via bidirectional attention at the concept-value level, generating local evidence and global relevance ($\boldsymbol{\alpha}_M$), which are gated to produce the initial node features $\mathbf{H}^{(0)}$. The \emph{PCG} models concept-value dependencies using PPMI-initialized edges refined through learnable message passing, resulting in the refined state $H^{(L)}$ that forms the interpretable bottleneck for diagnosis.
    }
    \label{fig:DCG-Net_architecture}
\end{figure*}

We propose \textbf{DCG-Net}, an end-to-end interpretable framework that unifies \emph{correlated concept reasoning} and \emph{cross-modal alignment} for transparent clinical diagnosis.
DCG-Net incorporates clinically structured priors, multimodal attention, and relational reasoning within a unified architecture. Unlike prior CBMs that treat concepts as independent or rely only on static similarity scores, DCG-Net jointly: (i) constructs canonical concept-value prototypes, (ii) performs bidirectional visual-textual grounding, and (iii) models dependencies between \emph{concept-value} pairs through a learnable graph, while preserving a strict concept bottleneck.
As illustrated in Fig.~\ref{fig:DCG-Net_architecture}, DCG-Net comprises three main components: a Concept Dictionary Encoder (Sec.~\ref{subsec:CDE}) for canonical concept-value embeddings, a DCA module (Sec.~\ref{subsec:DCA}) for bidirectional image-concept reasoning, and a PCG (Sec.~\ref{subsec:pcg}) for structured refinement of concept activations prior to diagnosis.

Let $I$ denote a medical image with diagnostic label $y \in \{1,\dots, C_y\}$, where $C_y$ denotes the number of diagnostic categories. We consider $K$ interpretable clinical concepts $\mathcal{C} = \{ c_1, \dots, c_K \}$, such as cell size. Each concept $c_k$ is associated with $M_k$ attribute values. For example, the concept \emph{cell size} may take values such as \emph{small} or \emph{large}. The total number of concept-value nodes is defined as
$ M = \sum_{k=1}^{K} M_k.$

\subsection{Concept Dictionary Encoder (CDE)}
\label{subsec:CDE}
Clinical datasets often contain heterogeneous or inconsistent terminology (\emph{e.g.}, `erythematous' vs.\ `red lesion'), which complicates the construction of stable concept representations. The CDE module addresses this by generating canonical textual prototypes for each concept value that serve as consistent semantic anchors for downstream reasoning.

For each concept $c_k$ with value $v_{k,m}$, we construct a set of prompts $\mathcal{U}_{k,m}$ by combining (i) the value name itself, (ii) its synonym set $\mathcal{S}(c_k)$, and (iii) a small collection of descriptive templates $\mathcal{T}$ that provide neutral clinical context. Formally, let
$\mathcal{S}(c_k)=\{s_{k,1},\dots,s_{k,n_k}\}$ and $\mathcal{T}=\{t_1,\dots,t_M\},$
where each template is a short clinical phrase constructed by inserting either the value name or a synonym into the placeholder of a template. This yields a diverse set of phrasing variations while preserving clinical meaning.
Each resulting prompt $u \in \mathcal{U}_{k,m}$ is encoded using the CLIP text encoder $f_{\text{text}}$ and $\ell_2$-normalized: \(\mathbf{e}(u)=f_{\text{text}}(u)/\|f_{\text{text}}(u)\|_2 \in \mathbb{R}^{d_t}\).

The canonical prototype for value $v_{k,m}$ is then defined as the average of its prompt embeddings:
\begin{equation}
\mathbf{t}_{k,m}=\frac{1}{|\mathcal{U}_{k,m}|}\sum_{u\in\mathcal{U}_{k,m}}\mathbf{e}(u).
\end{equation}
Stacking all prototypes gives the concept-value dictionary 
$T_M \in \mathbb{R}^{M\times d_t}$, which is projected into the visual embedding space via a learnable matrix $W_p$:
\begin{equation}
\tilde{T}_M = T_M W_p \in \mathbb{R}^{M \times d_v}.
\end{equation}
These projected prototypes serve as the fixed semantic anchors for cross-modal attention in subsequent modules, ensuring stable and clinically interpretable concept grounding.

\subsection{Dual Cross-Attention (DCA)}
\label{subsec:DCA}
Existing CLIP-based CBMs~\cite{wu2023medklip, oikarinen2023label,yang2023language, liu2025hybrid} rely on cosine similarity, which may not capture complex semantic alignment between image and text representations. Inspired by~\cite{lai2024carzero}, we design a DCA module to establish bidirectional alignment between the concept-value prototypes $\tilde{T}_M$ and visual tokens $V$. 

One branch identifies where each concept-value is supported in the image, while the other estimates its case relevance.
\subsubsection{Text-to-Image Attention (T2I)} 
A CLIP ViT-B/16 image encoder decomposes the image into patch-level visual tokens:
$V = [\mathbf{v}_0, \mathbf{v}_1, \dots, \mathbf{v}_P] \in \mathbb{R}^{(P+1) \times d_v}$, where \( \mathbf{v}_0 \) is the global [CLS] token representing global image context and \(\mathbf{v}_1, \dots, \mathbf{v}_P\) are patch tokens.

The \emph{local visual evidence} $\mathbf{F}_{\text{t2i}}$ for each concept value is obtained by querying the image patch tokens $V$ with the concept-value prototypes $\tilde{T}_M$ via multi-head attention (MHA), formulated as:
\begin{equation}
    \mathbf{F}_{\text{t2i}} = \mathrm{MHA}(\tilde{T}_M, V, V) \in \mathbb{R}^{M \times d_v}.
\end{equation}
Let $\mathbf{f}^{\text{t2i}}_m \in \mathbb{R}^{d_v}$ denote the visual evidence for concept value $m$ (e.g., `mitosis: present'), aggregated across the image patches.

\subsubsection{Image-to-Text Attention (I2T)} 
In parallel, this branch estimates the \emph{global relevance prior} for each value. The global [CLS] token $\mathbf{v}_0$ queries the concept-value prototypes $\tilde{T}_M$ to derive attention logits.
The global relevance score $\boldsymbol{\alpha}_M \in [0,1]^M$ for each node is then computed via a Sigmoid function on the logits:
\begin{equation}
    \boldsymbol{\alpha}_M = \mathrm{Sigmoid}\!\left(
        \frac{\mathbf{v}_0 \mathbf{K}_{\text{i2t}}^\top}{\tau}
    \right),
\end{equation}
where $\mathbf{K}_{\text{i2t}}$ are the projected keys of $\tilde{T}_M$ and $\tau$ is a temperature parameter. 
Let $\alpha_m$ denote the relevance score of node $m$,

\subsubsection{Feature Fusion}
\label{subsec:feature_fusion}
T2I determines \emph{where} a concept value is visually supported, whereas I2T determines \emph{how important} that value is for the case. Gating the former by the latter ensures that findings with stronger visual and clinical support are prioritized during downstream reasoning.
The initial node features
are obtained by gating the local visual evidence with the relevance scores: 
\begin{equation}
    \mathbf{h}^{(0)}_{m} = \alpha_{m} \, \mathbf{f}_{m}^{\text{t2i}},
\end{equation}
where $m \in \{1, \dots, M\}$ indexes the concept-value node. 
This direct fusion enables fine-grained cross-modal information to be used immediately for relational reasoning, removes the need for additional intermediate representations, and simplifies the interpretation of node features.

For each concept $k$, we compute a concept representation by mean pooling its $M_k$ associated value nodes,
$
\mathbf{c}_k = \operatorname{mean}_{m \in \mathcal{V}_k}(\mathbf{h}^{(0)}_m),
$
and then pass $\mathbf{c}_k$ through a concept-specific head $W^{\text{val}}_{k}$ to obtain the concept-level logit, formulated as
\begin{equation}
    u_k = W^{\text{val}}_{k}\, \mathbf{c}_k.
\end{equation}

\subsubsection{Alignment Loss}
To regularize the global relevance scores $\boldsymbol{\alpha}_M$, we introduce an alignment loss using binary cross-entropy (BCE), encouraging the model to assign higher relevance to concepts it predicts confidently.
Specially, the node-level relevance is first aggregated to the concept level by \( \alpha_k = \text{mean}_{m \in \mathcal{V}_k}(\alpha_m).\)
The alignment loss is defined as
\begin{equation}
\mathcal{L}_{\mathrm{align}} = \frac{1}{K}\sum_{k=1}^{K} \mathrm{BCE}(\alpha_k, \hat{t}_k),
\end{equation}
where $\hat{t}_k$ denotes the normalized confidence from the logit $u_k$.

\subsubsection{Concept Supervision}
\label{subsec:concept_loss}
To ensure that node features are semantically grounded, we supervise the classification logits derived from their representations.
Let $a_k \in \{0, \dots, M_k - 1\}$ denote the annotated concept-value label for concept $k$.
The resulting logits are trained using a cross-entropy (CE) loss $\mathcal{L}_{\text{concept}}$, encouraging accurate prediction of each concept’s values. Formally,
\begin{equation}
    \mathcal{L}_{\mathrm{concept}}
=
\frac{1}{K}
\sum_{k=1}^{K}
\mathrm{CE}(u_k, a_k).
\end{equation}

\subsubsection{Cross-Modal Consistency.}
To enforce coherence between the two cross-modal reasoning pathways, we introduce a consistency loss based on symmetrized Kullback-Leibler divergence (SKL) between the concept distributions inferred from T2I and I2T relevance estimation. Let $P^{\mathrm{T2I}}$ denote the normalized distribution over concepts obtained from T2I-derived concept scores, and let $P^{\mathrm{I2T}}$ denote the corresponding distribution obtained by normalizing the pooled relevance scores $\alpha_k$. The consistency loss is defined as 
\begin{equation}
\mathcal{L}_{\mathrm{cons}}
=
\mathrm{SKL}\!\left(P^{\mathrm{T2I}},\, P^{\mathrm{I2T}}\right),
\end{equation}
which penalizes disagreement between the two pathways and encourages consistent.

\subsection{Parametric Concept Graph (PCG)}
\label{subsec:pcg}
DCG-Net performs relational reasoning using the PCG, because clinical concepts exhibit statistical and semantic dependencies. It explicitly models and learns dependencies among all $M$ concept-value pairs, enabling refinement of initial evidence features $\mathbf{h}^{(0)}_{m}$ (see Sec.~\ref{subsec:feature_fusion}).

\subsubsection{Graph Definition and Rationale}
The PCG is a directed graph $\mathcal{G} = (\mathcal{V}, \mathcal{E})$ where each node $i \in \mathcal{V}$ corresponds to a specific concept-value pair $(k,m)$. The initial node features are $H^{(0)} = [\mathbf{h}^{(0)}_1;\dots;\mathbf{h}^{(0)}_M] \in \mathbb{R}^{M \times d_v}.$ 
This node definition provides reasoning at the concept-value level.

\subsubsection{Graph Parameterization and Adjacency Construction}
The adjacency matrix $\mathbf{A} \in \mathbb{R}^{M \times M}$ is derived from a structural prior combined with learnable parameters. We first construct a fixed PPMI (Positive Pointwise Mutual Information) prior $\mathbf{A}^{(0)}$ over concept-value co-occurrences using the training set, which serves as a structural bias toward clinically plausible associations. To refine this prior, we introduce a trainable score matrix $\mathbf{B} \in \mathbb{R}^{M \times M}$ and a binary structural mask $\mathbf{R}$ that removes invalid connections (e.g., self-loops and edges between values of the same concept). The unnormalized, non-negative edge weight $\tilde{A}_{ij}$ is defined as:
\begin{equation}
 \tilde{A}_{ij} = \mathrm{softplus}(B_{ij}) \cdot R_{ij} \cdot A^{(0)}_{ij}.
\end{equation}
This formulation ensures non-negativity and allows the model to learn contextually important edges while respecting fundamental clinical and structural rules encoded in $\mathbf{R}$ and $\mathbf{A}^{(0)}$. In particular, the PPMI prior anchors learning to empirically observed co-occurrences, which improves stability and interpretability compared to learning the full graph from scratch.

\subsubsection{Sparsity and Normalization}
To ensure computational stability and promote interpretability, the graph is sparsified and normalized. We first apply a row-wise $\mathrm{Top\textrm{-}k}$ operator to $\tilde{\mathbf{A}}$, retaining only the $k$ strongest outgoing connections for each node. Let $\bar{\mathbf{A}}$ be the resulting sparse adjacency matrix. This sparsity restricts message passing to neighborhoods reflecting clinically observed associations, rather than disseminating information across the entire graph. We then define the out-degree matrix $\mathbf{D} \in \mathbb{R}^{M \times M}$ as a diagonal matrix where $D_{ii}$ is the sum of outgoing edge weights from node $i$: \( D_{ii} = \sum_{j=1}^M \bar{A}_{ij} \). The final row-normalized stochastic adjacency matrix $\hat{\mathbf{A}}$ is then computed: \( \hat{\mathbf{A}} = \mathbf{D}^{-1}\bar{\mathbf{A}}\).
This row normalization ensures that the information flowing out of any node is correctly scaled for stable message passing and that weights can be interpreted as transition probabilities over concept-value relations. 

\subsubsection{Relational Propagation}
Given $\hat{\mathbf{A}}$ and initial node features $H^{(0)}$, the PCG performs $L$ layers of message passing, updating node features via a standard Graph Neural Network (GNN) layer:
\begin{equation}
 H^{(\ell+1)} = \sigma\!\left(H^{(\ell)} W^{(\ell)}_{\text{self}} + \hat{\mathbf{A}} H^{(\ell)} W^{(\ell)}_{\text{neigh}}\right),
\end{equation}
where $W^{(\ell)}_{\text{self}}, W^{(\ell)}_{\text{neigh}} \in \mathbb{R}^{d_v \times d_v}$ are learnable weight matrices, and $\sigma(\cdot)$ is a non-linear activation. After $L$ layers, we obtain the relationally refined node feature $H^{(L)} \in \mathbb{R}^{M \times d_v}$, which forms the interpretable bottleneck for final diagnosis. Because $H^{(L)}$ is produced by propagating information through a clinically grounded graph, its activations naturally encode both local visual evidence and global relational context.

\subsection{Diagnosis Prediction}
\label{sec:diaghead}
DCG-Net uses only the graph-refined concept-value state $H^{(L)}$ for diagnosis. We first aggregate $H^{(L)}$ across values for each concept $k$,
\begin{equation}
    \mathbf{c}^{\text{post}}_k = \text{mean}_{m \in \mathcal{V}_k}\!\left(H^{(L)}_m\right),
\end{equation}
and concatenate the resulting $K$ concept features into the final bottleneck vector
\begin{equation}
    \mathbf{z} = \text{vec}(\mathbf{c}^{\text{post}}_1; \dots; \mathbf{c}^{\text{post}}_K) \in \mathbb{R}^{K d_v}.
\end{equation}
The diagnosis logits $\mathbf{o}$ are computed via a linear classifier: 
\begin{equation}
\mathbf{o} = W_c \mathbf{z} + \mathbf{b}_c. 
\end{equation}
This design enforces that every prediction is strictly mediated by the learned concepts and their refined relationships.

The diagnosis head is supervised using a standard cross-entropy loss:
\begin{equation}
\mathcal{L}_{\mathrm{diag}} = \mathrm{CE}(\mathbf{o}, y),
\end{equation}
which enforces agreement between the predicted diagnosis and the clinical ground truth.

\subsection{Training Objective}
DCG-Net is trained end-to-end using a combination of diagnosis, concept grounding, and cross-modal regularization losses. The overall objective is
\begin{equation}
\mathcal{L}
=
\mathcal{L}_{\mathrm{align}}
+
\mathcal{L}_{\mathrm{concept}}
+
\mathcal{L}_{\mathrm{cons}}
+
\mathcal{L}_{\mathrm{diag}}.
\end{equation}
\section{Experimental Setup}
\label{sec:exp_setup}

\subsection{Datasets}
We evaluate DCG-Net on two medical image classification tasks. 
\textbf{Skin Lesion Diagnosis:} We use the Fitzpatrick17k dataset~\cite{groh2021evaluating}, covering benign, malignant, and non-neoplastic dermatologic conditions. Concept annotations are sourced from SkinCon~\cite{daneshjou2022skincon}, which provides 48 expert-defined clinical concepts; following~\cite{gao2024evidential,gao2025learning}, we use the 22 concepts represented by at least 50 images. Each image has a single diagnosis label and a multi-label set of binary concept annotations (22 concepts, present/absent).
\textbf{White Blood Cell (WBC) Morphology:} We use the Peripheral Blood Cell (PBC) dataset~\cite{acevedo2020dataset}. Concept-level supervision is obtained from the annotations in~\cite{tsutsui2023wbcatt}, which identify interpretable morphological attributes. Each image has a single diagnosis label and a set of categorical concepts, where each concept takes one of several mutually exclusive values.

\subsection{Key Implementation Details}
\subsubsection{Backbone and Fine-tuning}
We adopt CLIP ViT-B/16 as backbone. The text encoder is used to build the concept dictionary and is otherwise frozen. For the vision encoder, we freeze early transformer blocks and fine-tune only the last $L_v$ layers (typically $L_v = 2$ for SkinCon and $L_v = 4$ for WBC), together with the projection $W_p$, DCA, PCG, and classification heads. This strategy preserves the generalization ability of CLIP while allowing adaptation to the target medical domain and keeping training computationally manageable.

\subsubsection{Training Details}
We use AdamW with learning rate $5\times 10^{-5}$ and weight decay $5\times 10^{-3}$. A cosine schedule with 5\% warm-up is applied over the full training. We apply label smoothing for diagnosis, and use class-balanced weights for both diagnosis and concepts to mitigate label imbalance. For each dataset, we train DCG-Net with three different random seeds and select, for each run, the checkpoint with the highest validation macro-F1. We report test performance as mean and standard deviation across these three runs.

\subsubsection{Hardware and Implementation}
Experiments are implemented in \texttt{PyTorch 2.3.1} and executed on a Dell Precision~3660 workstation with Ubuntu~24.04 LTS, a 13th Gen Intel\textsuperscript{\textregistered} Core\texttrademark{} CPU, 32\,GB RAM, and an NVIDIA RTX~4090 GPU (24\,GB VRAM).

\subsection{Experimental Results and Discussion}
Table~\ref{tab:classification} illustrates that across both WBC and SkinCon benchmarks, DCG-Net improves diagnostic accuracy and concept prediction quality compared with existing concept-bottleneck approaches. In PBC dataset, DCG-Net achieves the highest diagnostic ACC while maintaining strong concept performance, confirming that the proposed dual cross-attention provides reliable visual-text grounding and that the concept-value graph effectively captures clinically meaningful dependencies. SkinCon results demonstrate similar trends: DCG-Net outperforms prior CBMs in both diagnosis and concept F1, suggesting that incorporating structured reasoning improves performance on fine-grained dermatologic attributes. 

\begin{table*}[htbp]
\caption{
Performance comparison on PBC and SkinCon classification tasks. 
\textbf{Bold} indicates the best results, and \underline{underlined} text denotes the second-best. 
Some compared results are directly reported from the cited paper~\cite{gao2025learning}.
For DCG-Net, concept metrics are computed on the pre-graph concept-value logits, which are directly supervised during training.
}
\renewcommand{\arraystretch}{1.2}
\begin{center}
\begin{tabular}{c| l| c| c| c| c}
\hline
\multirow{2}{*}{\textbf{Dataset}} & \multirow{2}{*}{\textbf{Method}} & 
\multicolumn{2}{c}{\textbf{Diagnosis Metrics}} & 
\multicolumn{2}{c}{\textbf{Concept Metrics}} \\
\cline{3-4}\cline{5-6}
 &  & Acc (\%) & F1 (\%) & Acc (\%) & F1 (\%) \\
\hline
\multirow{6}{*}{Fitzpatrick17k } 
& CBM~\cite{koh2020concept} (ICML 2020) & $78.37 \pm 0.70$ & $56.79 \pm 0.45$ & $78.42 \pm 0.50$ & $54.46 \pm 0.32$ \\
& CEM~\cite{espinosa2022concept} (NeurIPS 2022) & $77.07 \pm 1.44$ & $58.02 \pm 0.61$ & $78.96 \pm 0.50$ & $54.95 \pm 0.83$ \\
& ProbCBM~\cite{kim2023probabilistic} (ICML 2023) & $72.13 \pm 1.00$ & $54.03 \pm 0.15$ & $88.49 \pm 0.97$ & $57.01 \pm 1.52$ \\
& evi-CEM~\cite{gao2024evidential} (MICCA 2024) & $76.50 \pm 2.52$ & $58.55 \pm 2.27 $ & $90.33 \pm 0.37$ & \underline{$64.99 \pm 0.73$} \\
& Mica~\cite{bie2024mica}(AAAI 2024) & $75.63 \pm 1.07$ & $\mathbf{75.43 \pm 1.24}$ & $ 91.7 \pm --$ & $63.8 \pm --$ \\
& CGP~\cite{zhao2025concept} (MICCA 2025)  & $\underline{79.92 \pm --}$ & -- & $\mathbf{92.73 \pm --}$ & -- \\
& \textbf{DCG-Net (Ours)} & $\mathbf{83.44 \pm 0.31}$ & $\underline{72.13 \pm 0.99}$ & $\underline{92.18 \pm 0.05}$ & $\mathbf{71.05 \pm 0.02}$ \\
\hline

\multirow{7}{*}{PBC (WBC)} 
& CBM~\cite{koh2020concept} (ICML 2020) & $98.93 \pm 0.14$ & $98.44 \pm 0.22$ & $\underline{95.21 \pm 0.10}$ & $\underline{91.65 \pm 0.15}$ \\
& CEM~\cite{espinosa2022concept} (NeurIPS 2022) & $99.43 \pm 0.12$ & $99.23 \pm 0.18$ & $94.98 \pm 0.34$ & $90.86 \pm 0.69$ \\
& DCR~\cite{barbiero2023interpretable} (ICML 2023)& $98.93 \pm 0.32$ & $98.47 \pm 0.49$ & $92.79 \pm 0.43$ & $82.92 \pm 0.64$ \\
& WBCAtt~\cite{tsutsui2023wbcatt} (NeurIPS 2023)& - & - & - & $91.20 \pm 0.06$ \\
& CLR~\cite{gao2025learning} (MICCA 2025) & $98.67 \pm 0.25$ & $98.03 \pm 0.42$ & $\mathbf{95.32 \pm 0.26}$ & $90.51 \pm 0.73$ \\
& Align-CBM~\cite{pang2024integrating} (MICCA 2024) & $99.14 \pm 0.05$ & $99.25 \pm 0.52$ & $95.01 \pm 0.23$ & $89.95 \pm 0.59$ \\
& evi-CEM~\cite{gao2024evidential} (MICCA 2024)& $\underline{99.57 \pm 0.03}$ & $\underline{99.42 \pm 0.03}$ & $94.44 \pm 0.91$ & $89.18 \pm 2.09$ \\
& \textbf{DCG-Net (Ours)} & $\mathbf{99.76 \pm 0.07}$ & $\mathbf{99.66 \pm 0.06}$ & $94.52 \pm 0.03$ & $\mathbf{91.92 \pm 0.03}$ \\
\hline
\end{tabular}
\label{tab:classification}
\end{center}
\end{table*}

\begin{figure*}[htbp]
    \centering
    \includegraphics[width=\linewidth]{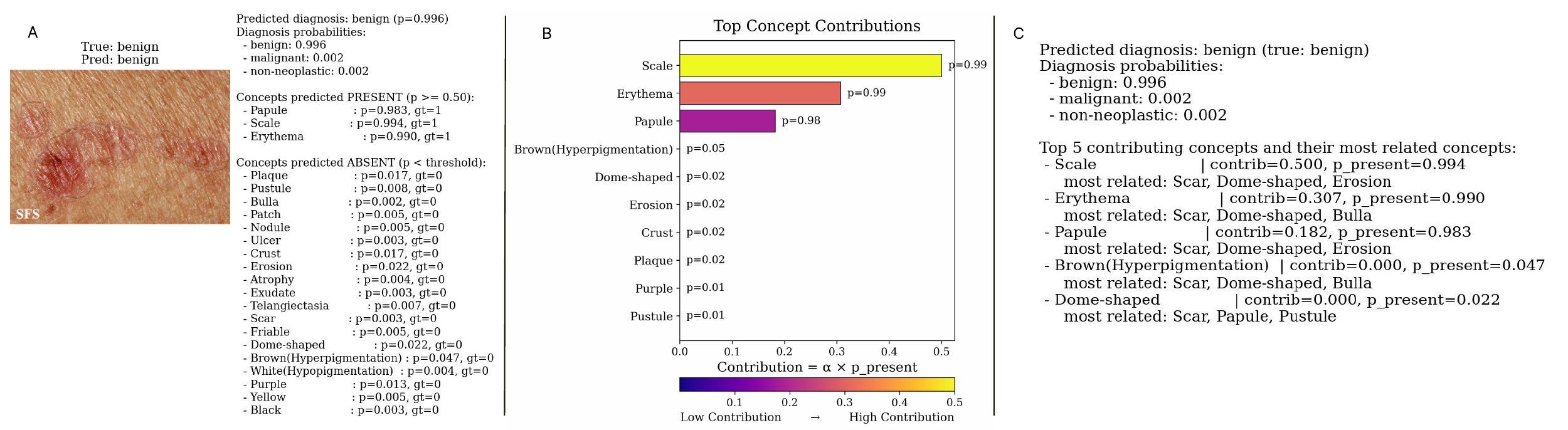}
    \caption{
    \textbf{Qualitative explanation of DCG-Net.}
    \textbf{(A)} Input image with predicted diagnosis, concept-value probabilities, and ground-truth labels. \textbf{(B)} Top concept contributions ($\alpha \times p_{\text{present}}$), showing which clinical findings most strongly support the model's diagnostic decision.
    \textbf{(C)} PCG-based relational reasoning, showing each top concept and its associated concept-value nodes, reflecting learned dependencies among concept-value pairs. 
    }
    \label{fig:qualitative}
\end{figure*}

\begin{table}[htbp]
\renewcommand{\arraystretch}{1.2}
\caption{Ablation for methodological justification.}
\begin{center}
\begin{tabular}{c|c|c}
\hline
\multirow{2}{*}{\textbf{Fitzpatrick17k}}&
\multicolumn{2}{|c}{\textbf{Performance Metrics}} \\
\cline{2-3}&
\textbf{\textit{Diagnosis (Acc/F1)}} &
\textbf{\textit{Concept (Acc/F1)}} \\
\hline
Full (ours) &
\textbf{83.44 / 72.13} &
\textbf{92.18 / 71.05} \\
\hline
w/o CDE &
82.51 / 71.36 &
91.95 / 68.87 \\
\hline
No PCG &
81.66 / 71.63 &
92.15 / 70.40 \\
\hline
\end{tabular}
\label{tab:min_ablate}
\end{center}
\end{table}

\subsection{Ablation Study}
Table~\ref{tab:min_ablate} further validates each component’s contribution: removing the CDE substantially degrades concept F1 due to weaker textual grounding; removing PCG reduces performance on correlated attributes by eliminating relational refinement. Together, these findings show that DCG-Net’s combination of concept normalization, bidirectional attention, and clinically informed graph propagation yields a structured and interpretable prediction process while achieving state-of-the-art performance on two distinct medical image domains.

\subsection{Qualitative Explanation}
Fig.~\ref{fig:qualitative} provides a qualitative analysis of DCG-Net’s interpretability on a Fitzpatrick17k~\cite{groh2021evaluating} sample by decomposing the diagnostic decision into concept prediction, contribution analysis, and relational reasoning. As shown in Fig.~\ref{fig:qualitative}(A), the model outputs concept-value probabilities alongside the predicted diagnosis, allowing the decision to be explicitly expressed via intermediate concepts and compared with annotated attributes. Fig.~\ref{fig:qualitative}(B) shows each concept’s relative contribution, computed as the product of concept relevance and predicted presence probability, highlighting the attributes most supportive of the prediction. Fig.~\ref{fig:qualitative}(C) further contextualizes these contributions using the parametric concept graph, linking top concepts to related concept-value nodes and revealing learned dependencies. Overall, DCG-Net produces interpretable diagnostic predictions through explicit concept modeling and graph-based reasoning.

\section{Conclusion}
\label{sec:conclusion}

We proposed \textbf{DCG-Net}, an end-to-end interpretable framework that unifies concept normalization, multimodal grounding, and relational reasoning for clinical image diagnosis. Through the \emph{CDE}, DCG-Net standardizes heterogeneous concept annotations; the \emph{DCA} module enables precise visual-textual alignment; and the \emph{PCG} captures clinically meaningful dependencies through data-informed relational refinement. This integrated design supports adaptive concept interaction, improves diagnostic accuracy, and preserves causal interpretability throughout the predictive pathway.
Experiments demonstrate strong performance, reliable concept prediction, and clinically aligned explanations that expose auditable intermediate reasoning.
Overall, DCG-Net represents progress toward interpretable, clinically informed diagnostic modeling by tightly coupling diagnostic predictions with explicit, clinically meaningful concepts and their learned relationships.

\bibliographystyle{IEEEbib}
\bibliography{reference}

\end{document}